\pdfoutput=1
\documentclass[11pt]{article}

\usepackage[]{ACL2023}

\usepackage{times}
\usepackage{latexsym}

\usepackage[T1]{fontenc}
\usepackage[utf8]{inputenc}

\usepackage{microtype}

\usepackage{inconsolata}
\usepackage{multirow}
\title{Gender Bias in Text: Labeled Datasets and Lexicons}

\author{Jad Doughman \\
  American University of Beirut \\
  Beirut, Lebanon \\
  \texttt{jad17@mail.aub.edu} \\\And
  Wael Khreich \\
  American University of Beirut \\
  Beirut, Lebanon \\
  \texttt{wk47@aub.edu.lb}}

\begin{document}
\maketitle
\begin{abstract}
Language has a profound impact on our thoughts, perceptions, and conceptions of gender roles. Gender-inclusive language is, therefore, a key tool to promote social inclusion and contribute to achieving gender equality. Consequently, detecting and mitigating gender bias in texts is instrumental in halting its propagation and societal implications. However, there is a lack of gender bias datasets and lexicons for automating the detection of gender bias using supervised and unsupervised machine learning (ML) and natural language processing (NLP) techniques. Therefore, the main contribution of this work is to publicly provide labeled datasets and exhaustive lexicons by collecting, annotating, and augmenting relevant sentences to facilitate the detection of gender bias in English text. Towards this end, we present an updated version of our previously proposed taxonomy by re-formalizing its structure, adding a new bias type, and mapping each bias subtype to an appropriate detection methodology. The released datasets and lexicons span multiple bias subtypes including: \textit{Generic He}, \textit{Generic She}, \textit{Explicit Marking of Sex}, and \textit{Gendered Neologisms}. We leveraged the use of word embedding models to further augment the collected lexicons.
\end{abstract}

\section{Introduction}
Given that language is the primary tool used to convey our perceptions, then any form of biased misrepresentation has the potential to change how an entity is portrayed in our minds. The source of bias in language can be traced to an androcentric worldview which was prevalent among 18th-century grammarians and was centered around the belief that: "human beings were to be considered male unless proven otherwise" \cite{Bodine1975}. Given that there is clear evidence of gender bias in most languages and its direct contribution to reinforcing and socializing sexist thinking \cite{Harris2017}, then there is a need to detect and highlight these manifestations in the ever-growing repertoire of textual content on the internet alongside printed writings such as educational textbooks.

Previously, most proposed solutions to detect gender bias in text were based on the frequency of gendered words and pronouns, in contrast, a feature-based approach would focus on capturing contextual and semantic queues in its classification process. Recently, Word Embedding (WE) and Contextual Word Embedding (CWE) models have become the predominant representation of text features, however, they are prone to capture biases inherited from training data. Despite the numerous attempts to debias these models, it was proven that these methods simply cover up systematic gender biases in word embeddings rather than removing them entirely \cite{Blodgett2020,Gonen2019}.

Hence, there is a need for labeled representative datasets and lexicons to train supervised learning models and employ lexicon-based approaches in an effort to automate the detection of gender bias in English text.  
Towards this end, we improve on a previously proposed gender bias taxonomy \cite{doughman-etal-2021-gender} by re-formalizing its structure, adding a new bias type, and mapping each bias subtype to an automated detection methodology.

Given that there is a clear lack of representative data with regard to gender bias, the main objective of this work is to collect and label datasets that encompass the various subtypes in the taxonomy. Several information retrieval and filtering methods were employed to collect representative sentences for each subtype of the taxonomy before we started labeling.

After having retrieved potentially biased sentences with a high recall, nine graduate-level annotators were assigned the task of labeling the sentences. The labeling process was guide-lined by the taxonomy and a few annotated examples of every bias subtype. To assess the clarity of the guidelines and the inter-rater reliability of the annotators, we computed Krippendorff’s alpha score across all annotations. In our case Krippendorff’s alpha was 0.75, confirming that the data or observations being evaluated are clear and unambiguous and that the raters have a shared understanding of the criteria being used to evaluate them. 

This work provides labeled datasets and exhaustive lexicons that can be leveraged by ML and NLP techniques for the automated detection of gender bias in texts. The released datasets and lexicons can be found here: \href{https://github.com/jaddoughman/Gender-Bias-Datasets-Lexicons}{Link}

\section{Related Work}

\subsection{Taxonomies}
Hitti et al. attempted to address bias at the sentence level and provided an initial categorization of gender bias types \cite{Hitti2019}. Doughman et al. developed a more comprehensive taxonomy to identify various types and subtypes of gender bias in English text \cite{doughman-etal-2021-gender}.  

\subsection{Datasets}
Due to the existing lack of a gender bias taxonomy in recent literature, most publicly available labeled datasets were centered around sexist statements, considering them the only form of gender bias. As shown below, almost all of the described datasets are addressing the two forms of sexist statements: benevolent sexism and hostile sexism. Hence, there is a need for well-representative datasets to detect the other overlooked forms of gender biases that are equally as detrimental.

\hfill\break\noindent\textbf{\citet{waseem-hovy-2016-hateful}} used various self-defined keywords to fetch tweets that are potentially sexist or racist by filtering the Twitter stream for two months \cite{waseem2016hateful}. The authors then labeled the data with the help of one outside annotator \cite{waseem2016hateful}. Additionally, they also annotated tweets that were neither sexist nor racist \cite{waseem2016hateful}. 

\hfill\break\noindent\textbf{\citet{jha-mamidi-2017-compliment}} augmented Waseem and Hovy’s dataset to include instances of benevolent sexism: sentences with a subjectively positive tone that implies that women are in need of special treatment and protection from men, and consequently furthering the stereotype of women as less capable \cite{jha2017does}. The authors collected data using terms, phrases, and hashtags that are "generally used when exhibiting benevolent sexism" and requested that three external annotators cross-validate the tweets to mitigate any annotator bias \cite{jha2017does}.

\hfill\break\noindent\textbf{\citet{fersini2018overview,fersini2020ami}} created the Automatic Misogyny Identification (AMI) competitions in Ibereval 2018 \cite{fersini2018overview} and Evalita 2020 \cite{fersini2020ami} provided datasets in English, Spanish, and Italian to detect misogynistic content, to classify misogynous behavior as well as to identify the target of a misogynous text.

\hfill\break\noindent\textbf{\citet{rodriguez2021overview}} collected a repertoire of popular sexist terms and expressions in both English and Spanish \cite{rodriguez2021overview}. The authors extracted the phrases and expressions from various tweets that women receive on a day-to-day basis on Twitter \cite{rodriguez2021overview}. The terms and expressions collected were commonly used to downplay and underestimate the role of women in our society \cite{rodriguez2021overview}.

\hfill\break\noindent\textbf{\citet{samory2021call}} gathered data from Twitter's Search API by utilizing the phrase “call me sexist(,) but" \cite{samory2021call}. To annotate their retrieved sentences using crowd-sourcing, they ran a pilot study and noticed that if interpreted as a disclaimer, annotators would have a tendency to presume that whatever follows “call me sexist(,) but" is automatically sexist \cite{samory2021call}. Consequently, they removed the given phrase for all annotation tasks (i.e., labeling requested that the annotator only label the remainder of each tweet (e.g. "Call me sexist, but please tell me why all women suck at driving." to "please tell me why all women suck at driving.") \cite{samory2021call}.

\hfill\break\noindent\textbf{\citet{chowdhury2019youtoo}} aggregated experiences of sexual abuse on Twitter using the "MeToo" hashtag to facilitate a better understanding of social media constructs and to bring about social change. \cite{chowdhury2019youtoo}. They released a comprehensive dataset and methodology for the detection of personal stories of sexual harassment on Twitter \cite{chowdhury2019youtoo}. Their work provided resources to clinicians, health practitioners, caregivers, and policymakers to identify communities at risk \cite{chowdhury2019youtoo}.

\begin{table*}[]
\begin{center}
 \begin{tabular}{l l l c} 
 \hline
  \textbf{Dataset} & \textbf{Reference} & \textbf{Labels} & \textbf{Size} \\
 \hline
 Waseem\&Hovy & \cite{waseem2016hateful}  & racism, sexism, neither & 16K\\ 
 Jha\&Mamidi & \cite{jha2017does} & benevolent, hostile, others & 22K\\
 AMI@Evalita & \cite{fersini2020ami} & misogynous, not misogynous & 10K\\
 AMI@IberEval & \cite{fersini2018overview} & misogynous, not misogynous & 8K\\
 EXIST@IberLEF & \cite{rodriguez2021overview} & sexist, not sexist & 11K\\
 “Call Me Sexist(,)" & \cite{samory2021call}  & (sexist, not sexist) + toxicity & 14K\\
 Chowdhury et al. & \cite{chowdhury2019youtoo} & recollection, not recollection & 5K\\
 \hline
\end{tabular}
\end{center}
\caption{Overview of datasets in related work}
\label{tab:related-work-datasets}
\end{table*}

\subsection{Lexicons}

\hfill\break\noindent\textbf{\citet{samory2021call}} worked on curating a selection of psychological scales, designed for measuring sexism and related constructs in individuals \cite{samory2021call}. The author's initial selection includes scales that were specifically designed to measure the construct of sexism or are frequently used to measure sexism in the social psychology literature \cite{samory2021call}. They further augmented their initial selection to include scales that were designed to measure constructs such as general attitudes towards men or women, egalitarianism, gender and sex role beliefs, stereotypical beliefs about men or women, attitudes towards feminism or gendered norms \cite{samory2021call}.

\hfill\break\noindent\textbf{\citet{bem1974measurement}} developed a sex-role inventory that considers masculinity and femininity as two distinct dimensions, thereby halting the ability to characterize a person as masculine, feminine, or "androgynous" as a function of the difference between her or his endorsement of masculine and feminine personality traits \cite{bem1974measurement}.

\hfill\break\noindent\textbf{\citet{spence1974personal}} augmented the conceptual analysis of the PAQ by appending a larger variety of self-reported measures. Additionally, the base includes data from an entirely new domain of personality measurement-observer ratings \cite{spence1974personal}. 

% \begin{table}[htbp]
% \caption{Overview of lexicons in related work} 
% \begin{center}
%  \begin{tabular}{l c l} 
%  \hline
%   \textbf{Lexicon} & \textbf{Type} \\
%  \hline
%  Sexism Psychological Scales & scales\\ 
%  Bem Sex Role Inventory (BSRI)  & inventory \\ 
%  Personal Attributes Questionnaire & questionnaire\\ 
%  \hline
% \end{tabular}
% \label{tab:related-work-lexicons}
% \end{center}
% \end{table}

\section{Improved Gender Bias Taxonomy} \label{taxonomy}
The first step of detecting biased language is to categorize the various forms of that bias while carefully maintaining clear segregation between the resultant groups. This section presents an updated version of our previously proposed taxonomy by re-formalizing its structure, adding a new bias type (stereotyping), and mapping each bias subtype to an appropriate detection methodology. Table \ref{linkage} provides an overview of the taxonomy, with one example pertaining to each subtype. The table also maps each bias subtype to its most practical detection methodology (supervised learning or lexicon-based).

\subsection{Stereotyping Bias}
Stereotypes have been defined in a variety of ways within the literature, however, this paper adopts the standard viewpoint that stereotypes are beliefs about the characteristics, attributes, and behaviors of members of certain groups \cite{hilton1996stereotypes}. We also segregate stereotyping into two types: societal and behavioral.

\subsubsection{Societal Stereotype} Societal stereotypes depict traditional gender roles that reflect social norms. \cite{tajfel2010social,Hitti2019}. Below are a few examples that depict the concept of societal stereotypes: 
\begin{itemize}
    \item Senators need their wives to support them throughout their campaigns.
    \item The event was kid-friendly for all the mothers working in the company.
\end{itemize}
 
\subsubsection{Behavioural Stereotype} Behavioural sentences contain attributes and traits that are being generalized onto a person or gender. Below are a few examples that depict the concept of behavioral stereotypes: 
 \begin{itemize}
     \item All boys are aggressive.
     \item Mary must love dolls because all girls like playing with them.
 \end{itemize}
 
% hbt!
\begin{table*}[] 
\begin{tabular}{ p{0.22\linewidth}  p{0.25\linewidth}  p{0.28\linewidth}  p{0.15\linewidth} }
\hline
\textbf{Bias Type}                       & \textbf{Bias Subtype}           & \textbf{Example}                                         & \textbf{Methodology}     \\ \hline
                                  \\  & Generic He                 & A programmer must carry his laptop with him to work.              & Supervised Learning    \\  %\cline{2-4} 
\multirow{-1}{*}{\textbf{Generic Pronouns}}   \\   & Generic She                & A nurse should ensure that she gets adequate rest.            & Supervised Learning    \\ \\ \hline
                                 \\   & Societal Stereotypes         & Senators need their wives to support them throughout their campaign. & Supervised Learning    \\ \\ %\cline{2-4} 
\multirow{-3}{*}{\textbf{Stereotyping Bias}}  & Behavioural Stereotypes       & The event was kid-friendly for all the mothers working in the company.  & Supervised Learning    \\ \\ \hline
                              \\     & Hostile Sexism             & Women are incompetent at work.                             & Supervised Learning     \\ \\ %\cline{2-4} 
\multirow{-3}{*}{\textbf{Sexism}}   & Benevolent Sexism          & They’re probably surprised at how smart you are, for a girl.   & Supervised Learning  \\ \\ \hline
                              \\     & Explicit Marking of Sex    & Chairman, Businessman, Manpower, Cameraman                    & Lexicon-Based  \\ %\cline{2-4} 
\multirow{-1}{*}{\textbf{Exclusionary Terms}}  \\     & Gendered Neologisms    & Man-bread, Man-sip, Man-tini                                 & Lexicon-Based   \\\\ \hline
                                  \\ & Metaphors                  & ``Cookie": lovely woman.                                 & Supervised Learning       \\ %\cline{2-4} 
\multirow{-1}{*}{\textbf{Semantic Bias}}  \\       & Old Sayings                & A woman’s tongue three inches long can kill a man six feet high.    & Supervised Learning    \\  \\ \hline
\end{tabular}
\caption{Overview of the taxonomy and link to detection methodology}
\label{linkage}
\end{table*}

\section{Generic Pronoun Datasets}
Detecting any form of gender bias in English text in a supervised learning fashion is contingent on a labeled dataset that conforms with its linguistic requirements. To date, no labeled datasets have been publicly released pertaining to generic pronouns. Towards this end, the below sections describe the automated data retrieval methodology utilized to retrieve generic pronoun sentences, the annotation tool/process utilized to label the retrieved sentences, and the inter-rater agreement of the contributors. 

\subsection{Automated Data Retrieval} \label{generic-pronoun-retrieval}
A pronoun typically follows the sex of its referent. However, when the referent is sex-indefinite, the pronoun becomes generic since it would be generalizing the sex of the pronoun onto the gender-neutral entity it's referencing. The most notable form of a generic pronoun sentence occurs when a pronoun's referent is a sex-indefinite occupation. Taking the below examples, we notice that a pattern prevails. 
\begin{itemize}
    \item \textbf{S1:} A \textit{programmer} must always carry \textit{his} laptop to work.
    \item \textbf{S2:} A \textit{nurse} should ensure that \textit{she} gets adequate rest.
    \item \textbf{Pattern:} "A {\textit{occupation}} * {\textit{pronoun}}"
\end{itemize}
When a pronoun refers to an occupation rather than a sex-definite person (subject), it becomes generic. In an effort to retrieve potential generic pronoun sentences in an automated fashion, we combined the above linguistic pattern with advanced information retrieval queries. We used first initialized a list of around 1156 occupations and all possible gendered pronouns. For each combination, we used Google Search API to return and store the results of this template query ("A {\textit{occupation}} * {\textit{pronoun}}"). Having applied the above pattern, we were able to retrieve biased sentences with a high recall score. However, we did notice that there were instances where the above pattern occurs, but the pronoun does not end up being generic. Taking the below examples, it is clear that a sentence could contain both a pronoun and an occupation, but the pronoun's referent wouldn't be the occupation but rather a sex-definite person. 
\begin{itemize}
    \item \textbf{Biased:} "A \textit{programmer} must always carry \textit{his} laptop to work" 
    \item \textbf{Not Biased:} "John, a \textit{programmer}, always carries \textit{his} laptop to work" 
\end{itemize}
In order to increase the recall of positive (biased) instances in our retrieval process, we categorized the retrieved sentences into three types: declarative, imperative, and interrogative. We concluded that when the agreed-upon pattern is formalized in an interrogative manner (question), it most frequently happens to be biased. This case is especially valid in question-answering platforms since the questioner would not be referencing a person, but rather asking a question in a general manner. The below examples illustrate our hypothesis:
\begin{itemize}
    \item "How often does a programmer update his skills?"
    \item "Can you identify a programmer based on his code?"
\end{itemize}
As shown above, general interrogative sentences typically reduce the chance of a sex-definite subject occurring, which results in more biased sentences retrieved. Alternatively, declarative sentences typically contain a vague reference or a reference to a known subject in previous sentences. However, given that we are currently solely retrieving and labeling one distinct sentence at a time, a multi-sentence reference is problematic in associating a pronoun from one sentence to a potentially sex-definite subject in another sentence. For future work, we aim to retrieve and label paragraphs rather than sentences and integrate co-reference resolution to minimize any ambiguity regarding the sex of a pronoun's referent. We will also aim to investigate the effect of generic pronoun questions on the bias-ness of the answer-er. 

Based on the above findings, we focused on automating the retrieval of sentences that conform with our pattern for positive instances. The total number of sentences retrieved is 700. The dataset spans 29 occupations with at least 20 sentences per occupation. The retrieved sentences, alongside their annotations, were augmented to reach 3,510 sentences as detailed in Section \ref{dataset-augmentation}.

\subsection{Annotation Process}
After having retrieved 700 potentially biased sentences following the patterns described in Section \ref{generic-pronoun-retrieval}, we loaded the sentences into INCEpTION, a semantic annotation tool used for concept linking, fact linking, and knowledge base population \cite{klie2018inception}. The total number of contributors tasked with labeling the sentences is nine. All contributors are graduate-level university students with extensive experience regarding gender bias. Additionally, most annotators are familiar with the gender bias taxonomy described in Section \ref{taxonomy}, which further enhanced their understanding of biased and non-biased statements. 

The contributors were tasked with labeling each sentence as biased or not. If the sentence was biased, the annotators were also asked to highlight the generic pronoun and its sex-indefinite referent (occupation). If the sentence was not biased, the annotators were asked to highlight the non-generic pronoun and its sex-definite referent (subject). To ensure that the annotators were well-equipped to differentiate between the required classes, a guideline of around 10 sentences was presented. Furthermore, a few golden standard sentences were randomly inserted to evaluate the contributor's understanding of the labeling process. Table \ref{tab:annotation} illustrates a few examples.

\begin{table*}[htbp]
\begin{center}
  \begin{tabular}{p{0.8\textwidth} p{0.14\textwidth}}
    \hline
    \textbf{Sentence} & \textbf{Label}\\
    \hline
    A \textbf{programmer} must always carry \textbf{his} laptop to work. & Biased \\
    How do I tell if a \textbf{pastor} is turning \textbf{his} congregation into a cult? & Biased \\
    \textbf{Jennie} is a rapper, \textbf{her} voice is suited for rapping. & Not Biased \\
    He always had a \textbf{dietitian} in \textbf{his} course to help him. & Not Biased  \\
    Can you judge a \textbf{nurse's} professionalism by \textbf{his/her} demeanor in the nurse station? & Avoiding Bias \\
  \hline
\end{tabular}
\end{center}
\caption{Overview of annotation process}
\label{tab:annotation}
\end{table*}

As shown in Table \ref{tab:annotation}, there are instances where the pronoun is referencing a sex-indefinite occupation but the sentence is not biased. We call these cases: "Avoiding Bias" since we consider that the writer is aware of the gender bias and is avoiding it by replacing a generic pronoun with "his/her". For future work, replacing a generic pronoun with "his/her" or "her/his" could be a viable gender bias mitigation technique.  

\subsection{Krippendorff's Alpha-Reliability}
Krippendorff’s alpha is a reliability coefficient to measure the agreement among multiple annotators \cite{krippendorff2011computing}. The significance of the reliability of a rater stems from the fact that it signifies the degree to which the data collected in the study are correct representations of the variables measured. Thus, inter-rater reliability is defined as the extent to which data collectors (raters) award the same score to the same variable. In our case, Krippendorff’s alpha was 0.75 which suggests that the data or observations being evaluated are clear and unambiguous and that the raters have a shared understanding of the criteria being used to evaluate them. 

\subsection{Dataset Augmentation} \label{dataset-augmentation}
To augment our dataset based on our initial annotations, we generated multiple variants of the same sentence by solely altering its pronoun. Provided that appending the opposite pronoun "her/his" or "his/her" would negate a sentence's bias, we replaced every generic pronoun, such as "her", with a negation which resulted in a non-biased variant of the primary sentence. Alternatively, replacing "his/her" or "her/his" with "his" in one sentence and "her" in another sentence resulted in two new biased sentences from one unbiased one. The examples below illustrate the process of generating two new biased sentences from one unbiased sentence using our proposed technique:
\begin{itemize}
    \item \textbf{Original Sentence:} How often does a programmer update \textbf{his/her} skills? 
    \item \textbf{Augmented Sentence \#1:} How often does a programmer update \textbf{his} skills?
    \item \textbf{Augmented Sentence \#2:} How often does a programmer update \textbf{her} skills?
\end{itemize}
The integrity of the annotations is preserved since the augmentation process solely altered one token in each sentence and thus did not change its overall meaning. Augmenting the dataset resulted in 2,400 additional sentences, which will be publicly released. 

\section{Exclusionary Terms Lexicon}
Exclusionary terms occur when an unknown gender-neutral entity is referred to using gender-exclusive term(s). One example is adding the gender-exclusive sub-word (e.g. "man") onto a gender-neutral occupational term (e.g. "Police"), resulting in "Policeman". The resultant biased word implies that all police officers are men, which excludes women. The reverse, concatenating "woman" with "Police" resulting in "Policewoman", is also applicable since it would be implying that all police officers are women. The presence of exclusionary terms in language has proven to have various negative societal implications. For instance, sex-biased wording affects a person’s perception of a career’s attractiveness \cite{briere1983sex}. Consequently, countries that adopt a gendered language tend to have disproportionate labor force participation \cite{gay2013grammatical}. Furthermore, the presence of gender bias in the language used by parents and in school textbooks may cause children to develop sexist perceptions and behaviors towards other children of the opposite gender and deepens the problematic outcomes of gender inequalities in society \cite{waxman2013building}. Therefore, the aim of the below sections is to provide a list of terms that are exclusionary, in hopes of halting their propagation and subsequently their societal implications.

\subsection{Explicit Marking of Sex Lexicon}  \label{lexicon-augmentation}
The section describes the source of the initial lexicon and the NLP techniques utilized to augment the word list. The initial explicit marking of sex lexicon was curated from a guideline report published by the United Nations Economic and Social Commission for West Asia (UNESCWA) "Gender Sensitive Language" \cite{unescwa}. They provided a list of violating terms and proposed a correction for each one. However, the list only spans 86 words and is not sufficient enough to cover all exclusionary terms in the English language. To augment the initial lexicon, we leveraged a word embedding model's ability to associate terms that have similar meanings using the cosine similarity of their vector representations. We started by loading various pre-trained Word2Vec \cite{mikolov2013distributed} models that possess a large vocabulary size. We then computed the cosine similarity of each lexicon word against all other words in a model's vocabulary. The results were ranked in descending order of cosine similarity values, essentially pinning the most similar word vectors on top. We then selected each of the top 100 most similar word vectors and appended them into a set of unique similar words. This resulted in a set of 8,600 distinct words that are potentially exclusionary. Below is an example of the top-5 most similar word vectors to "Salesman":

Although some word vectors are close in the embedding space to the original exclusionary term, however, they are not necessarily biased since they aren't unknown gender-neutral entities being referred to using a gender-exclusive term. To filter out non-exclusionary terms, we kept the tokens that contain a gender-exclusive sub-string (e.g. "man"). For example, we would filter out words such as "feminine" and retain words such as the word "womanly". This rule proved to be effective in filtering out words that do not contain sub-strings that explicitly mark a certain sex. To conclude, for a word to be appended to this lexicon, it has to be close in the embedding space to a valid lexicon exclusionary word, contains a "man" sub-string, and is validated by an annotator. Overall the number of explicit marking of sex terms increased to 145.

\begin{table*}[htbp]
  \begin{center}
  \begin{tabular}{p{0.1\textwidth} p{0.58\textwidth} cc}
    \hline
    \textbf{Word} & \textbf{Definition} & \textbf{Up Votes} & \textbf{Down Votes} \\
    \hline
    Manboobs & Name given to a Male's breasts when they grow to abnormally large size. Manboobs are common on the heavier sized males, and are not to be mistaken for a normal female's breasts. & 167 & 67 \\
    \hline
    Manpons & Similar to the feminine tampon, the masculine "manpon" is used for the reduction of sweat between the cheeks of the buttocks, placed firmly between the cheeks in times of high pressure, stress, or sweat-causing situations. & 338 & 201\\
    \hline
    Man-sip & A man sized sip of a beer or drink, one can finish a beer in 4 or 5 Man-sips For a female or light weight it borders on chugging the drink, but for a man it is merely a sip. & 14 & 1\\
  \hline
\end{tabular}
\end{center}
\caption{Urban dictionary samples}
\label{tab:urban-dictionary}
\end{table*}

\subsection{Gendered Neologisms}
Neologisms are newly coined terms that are in the process of mainstream adoption; however, they have not yet been entirely recognized. Gender-based neologisms are therefore gendered exclusionary coinages with underlying biased tendencies \cite{foubert2018gender}. They are analogous to explicit marking of sex terms, in which they're both exclusionary, but differ in their adoption rate. Explicit marking of sex terms is more commonly used and accepted terms such as "Policeman" and "Businessman" while gender-based neologisms are newly coined and are in the process of mainstream adoption such as "Man-tini" and "Man-bread". Thus, the significance of detecting and mitigating gender-based neologisms is critical in halting its propagation and ability to become widely adopted and integrated into the English language. To this end, this section describes the process of curating and filtering gender-based neologism terms from the Urban Dictionary \cite{peckham2009urban}.

\subsubsection{Urban Dictionary} Urban Dictionary (UD) is a popular online slang dictionary that is built by the collaboration of contributing end-users, allowing people who are not trained lexicographers to engage in the actual making of dictionaries \cite{peckham2009urban}. However, because UD is an open-source platform, any internet user may submit a new dictionary term entry that they feel is or should be used in a mainstream way \cite{peckham2009urban}. This becomes troublesome when the newly formed phrases supplied are prejudiced, and their acceptance might be destructive to society. Table \ref{tab:urban-dictionary} illustrates a few examples of dictionary word entries on UD that have exclusionary and stereotypical tendencies. 

In an effort to detect and mitigate the adoption of such terms, the aim of the below section is to provide a means of finding such exclusionary terms among the UD by filtering them through specific sub-strings and up-vote counts. 

\subsubsection{Filtering Technique}
Given that the UD currently stands at more than 2 million words in total, it wasn't feasible for us to manually go through and label each word as biased or not. To accurately select newly coined exclusionary terms from UD, a two-step process is presented. Firstly, we compartmentalized the dictionary by filtering out all the words that don't contain a gender-exclusive sub-word (e.g. "man"). This left us with around 25,000 newly coined terms that are potentially exclusionary. To trim the dictionary even further, we filtered out words that have less than 100 up-votes to distinguish between terms that are accepted by the community and are on the brink of mainstream adoption compared to words that the community itself is against its use. This step further reduced the dictionary size to around 2,500 potentially biased terms. We finally manually labeled the remaining sentences as exclusionary based on the author's definition of the term. The final lexicon spans around 500 newly coined biased terms. We hope that our work provides a means for the technical community to detect and mitigate the use of such terms to halt their propagation and subsequent adoption. 

\section{Conclusion}
In conclusion, the fundamental contribution of this work is to offer labeled datasets and exhaustive lexicons by collecting, annotating, and augmenting representative sentences. This work also offers insight into the automated data retrieval and annotation methodologies utilized to fetch and label the retrieved sentences. In future work, we will address the issue of pronoun resolution by considering surrounding sentences or entire paragraphs. We will also aim to further augment our datasets and lexicons to expand their coverage to the remaining bias types. We hope that the labeled datasets and lexicons, backed by our improved taxonomy, can pave the way for the technical community to detect and mitigate gender bias in English texts using ML and NLP techniques.

\section{Acknowledgment}
This project was generously funded by the University Research Board (URB) - Faculty Research Grants Program (FRGP) at the American University of Beirut. Project \# 3285.

% Entries for the entire Anthology, followed by custom entries
\bibliography{anthology,custom}

\begin{thebibliography}{28}
\expandafter\ifx\csname natexlab\endcsname\relax\def\natexlab#1{#1}\fi

\bibitem[{Bem(1974)}]{bem1974measurement}
Sandra~L Bem. 1974.
\newblock The measurement of psychological androgyny.
\newblock \emph{Journal of consulting and clinical psychology}, 42(2):155.

\bibitem[{Blodgett et~al.(2020)Blodgett, Barocas, Daum{\'{e}}, and
  Wallach}]{Blodgett2020}
Su~Lin Blodgett, Solon Barocas, Hal Daum{\'{e}}, and Hanna Wallach. 2020.
\newblock \href {http://arxiv.org/abs/2005.14050} {{Language (Technology) is
  Power: A Critical Survey of "Bias" in NLP}}.

\bibitem[{Bodine(1975)}]{Bodine1975}
Ann Bodine. 1975.
\newblock \href {https://doi.org/10.1017/S0047404500004607} {Androcentrism in
  prescriptive grammar: singular ‘they’, sex-indefinite ‘he’, and ‘he
  or she’}.
\newblock \emph{Language in Society}, 4:129 -- 146.

\bibitem[{Briere and Lanktree(1983)}]{briere1983sex}
John Briere and Cheryl Lanktree. 1983.
\newblock Sex-role related effects of sex bias in language.
\newblock \emph{Sex roles}, 9(5):625--632.

\bibitem[{Chowdhury et~al.(2019)Chowdhury, Sawhney, Shah, and
  Mahata}]{chowdhury2019youtoo}
Arijit~Ghosh Chowdhury, Ramit Sawhney, Rajiv Shah, and Debanjan Mahata. 2019.
\newblock \# youtoo? detection of personal recollections of sexual harassment
  on social media.
\newblock In \emph{Proceedings of the 57th Annual Meeting of the Association
  for Computational Linguistics}, pages 2527--2537.

\bibitem[{Doughman et~al.(2021)Doughman, Khreich, El~Gharib, Wiss, and
  Berjawi}]{doughman-etal-2021-gender}
Jad Doughman, Wael Khreich, Maya El~Gharib, Maha Wiss, and Zahraa Berjawi.
  2021.
\newblock \href {https://doi.org/10.18653/v1/2021.gebnlp-1.5} {Gender bias in
  text: Origin, taxonomy, and implications}.
\newblock In \emph{Proceedings of the 3rd Workshop on Gender Bias in Natural
  Language Processing}, pages 34--44, Online. Association for Computational
  Linguistics.

\bibitem[{Fersini et~al.(2020)Fersini, Nozza, and Rosso}]{fersini2020ami}
Elisabetta Fersini, Debora Nozza, and Paolo Rosso. 2020.
\newblock Ami@ evalita2020: Automatic misogyny identification.
\newblock In \emph{EVALITA}.

\bibitem[{Fersini et~al.(2018)Fersini, Rosso, and
  Anzovino}]{fersini2018overview}
Elisabetta Fersini, Paolo Rosso, and Maria Anzovino. 2018.
\newblock Overview of the task on automatic misogyny identification at ibereval
  2018.
\newblock \emph{IberEval@ SEPLN}, 2150:214--228.

\bibitem[{Foubert and Lemmens(2018)}]{foubert2018gender}
Oc{\'e}ane Foubert and Maarten Lemmens. 2018.
\newblock Gender-biased neologisms: the case of man-x.
\newblock \emph{Lexis. Journal in English Lexicology}, (12).

\bibitem[{Gay et~al.(2013)Gay, Santacreu-Vasut, and
  Shoham}]{gay2013grammatical}
Victor Gay, Estefania Santacreu-Vasut, and Amir Shoham. 2013.
\newblock The grammatical origins of gender roles.
\newblock \emph{Berkeley Economic History Laboratory Working Paper}, 3.

\bibitem[{Gonen and Goldberg(2019)}]{Gonen2019}
Hila Gonen and Yoav Goldberg. 2019.
\newblock \href {http://arxiv.org/abs/1903.03862} {{Lipstick on a Pig:
  Debiasing Methods Cover up Systematic Gender Biases in Word Embeddings But do
  not Remove Them}}.
\newblock pages 1--6.

\bibitem[{Harris et~al.(2017)Harris, Blencowe, and Telem}]{Harris2017}
Chelsea Harris, Natalie Blencowe, and Dana Telem. 2017.
\newblock \href {https://doi.org/10.1097/SLA.0000000000002505} {What is in a
  pronoun?: Why gender-fair language matters}.
\newblock \emph{Annals of Surgery}, 266:1.

\bibitem[{Hayek(2021)}]{unescwa}
Nouhad Hayek. 2021.
\newblock \href
  {https://www.unescwa.org/sites/default/files/services/doc/guidelines_gender-sensitive_language_e-a.pdf}
  {Gender-sensitive language}.

\bibitem[{Hilton and Von~Hippel(1996)}]{hilton1996stereotypes}
James~L Hilton and William Von~Hippel. 1996.
\newblock Stereotypes.
\newblock \emph{Annual review of psychology}, 47(1):237--271.

\bibitem[{Hitti et~al.(2019)Hitti, Jang, Moreno, and Pelletier}]{Hitti2019}
Yasmeen Hitti, Eunbee Jang, Ines Moreno, and Carolyne Pelletier. 2019.
\newblock \href {https://doi.org/10.18653/v1/w19-3802} {{Proposed Taxonomy for
  Gender Bias in Text; A Filtering Methodology for the Gender Generalization
  Subtype}}.
\newblock pages 8--17.

\bibitem[{Jha and Mamidi(2017{\natexlab{a}})}]{jha-mamidi-2017-compliment}
Akshita Jha and Radhika Mamidi. 2017{\natexlab{a}}.
\newblock \href {https://doi.org/10.18653/v1/W17-2902} {When does a compliment
  become sexist? analysis and classification of ambivalent sexism using twitter
  data}.
\newblock In \emph{Proceedings of the Second Workshop on {NLP} and
  Computational Social Science}, pages 7--16, Vancouver, Canada. Association
  for Computational Linguistics.

\bibitem[{Jha and Mamidi(2017{\natexlab{b}})}]{jha2017does}
Akshita Jha and Radhika Mamidi. 2017{\natexlab{b}}.
\newblock When does a compliment become sexist? analysis and classification of
  ambivalent sexism using twitter data.
\newblock In \emph{Proceedings of the second workshop on NLP and computational
  social science}, pages 7--16.

\bibitem[{Klie et~al.(2018)Klie, Bugert, Boullosa, de~Castilho, and
  Gurevych}]{klie2018inception}
Jan-Christoph Klie, Michael Bugert, Beto Boullosa, Richard~Eckart de~Castilho,
  and Iryna Gurevych. 2018.
\newblock The inception platform: Machine-assisted and knowledge-oriented
  interactive annotation.
\newblock In \emph{Proceedings of the 27th International Conference on
  Computational Linguistics: System Demonstrations}, pages 5--9.

\bibitem[{Krippendorff(2011)}]{krippendorff2011computing}
Klaus Krippendorff. 2011.
\newblock Computing krippendorff's alpha-reliability.

\bibitem[{Mikolov et~al.(2013)Mikolov, Sutskever, Chen, Corrado, and
  Dean}]{mikolov2013distributed}
Tomas Mikolov, Ilya Sutskever, Kai Chen, Greg~S Corrado, and Jeff Dean. 2013.
\newblock Distributed representations of words and phrases and their
  compositionality.
\newblock In \emph{Advances in neural information processing systems}, pages
  3111--3119.

\bibitem[{Peckham et~al.(2009)}]{peckham2009urban}
Aaron Peckham et~al. 2009.
\newblock \emph{Urban dictionary: Fularious street slang defined}, volume~1.
\newblock Andrews McMeel Publishing.

\bibitem[{Rodr{\'\i}guez-S{\'a}nchez et~al.(2021)Rodr{\'\i}guez-S{\'a}nchez,
  Carrillo-de Albornoz, Plaza, Gonzalo, Rosso, Comet, and
  Donoso}]{rodriguez2021overview}
Francisco Rodr{\'\i}guez-S{\'a}nchez, Jorge Carrillo-de Albornoz, Laura Plaza,
  Julio Gonzalo, Paolo Rosso, Miriam Comet, and Trinidad Donoso. 2021.
\newblock Overview of exist 2021: sexism identification in social networks.
\newblock \emph{Procesamiento del Lenguaje Natural}, 67:195--207.

\bibitem[{Samory et~al.(2021)Samory, Sen, Kohne, Fl{\"o}ck, and
  Wagner}]{samory2021call}
Mattia Samory, Indira Sen, Julian Kohne, Fabian Fl{\"o}ck, and Claudia Wagner.
  2021.
\newblock Call me sexist, but…: Revisiting sexism detection using
  psychological scales and adversarial samples.
\newblock In \emph{Intl AAAI Conf. Web and Social Media}, pages 573--584.

\bibitem[{Spence et~al.(1974)Spence, Helmreich, and Stapp}]{spence1974personal}
Janet~T Spence, Robert~L Helmreich, and Joy Stapp. 1974.
\newblock \emph{The Personal Attributes Questionnaire: A measure of sex role
  stereotypes and masculinity-femininity}.
\newblock University of Texas.

\bibitem[{Tajfel(2010)}]{tajfel2010social}
Henri Tajfel. 2010.
\newblock Social stereotypes and social groups.

\bibitem[{Waseem and Hovy(2016{\natexlab{a}})}]{waseem-hovy-2016-hateful}
Zeerak Waseem and Dirk Hovy. 2016{\natexlab{a}}.
\newblock \href {https://doi.org/10.18653/v1/N16-2013} {Hateful symbols or
  hateful people? predictive features for hate speech detection on {T}witter}.
\newblock In \emph{Proceedings of the {NAACL} Student Research Workshop}, pages
  88--93, San Diego, California. Association for Computational Linguistics.

\bibitem[{Waseem and Hovy(2016{\natexlab{b}})}]{waseem2016hateful}
Zeerak Waseem and Dirk Hovy. 2016{\natexlab{b}}.
\newblock Hateful symbols or hateful people? predictive features for hate
  speech detection on twitter.
\newblock In \emph{Proceedings of the NAACL student research workshop}, pages
  88--93.

\bibitem[{Waxman(2013)}]{waxman2013building}
Sandra~R Waxman. 2013.
\newblock Building a better bridge.
\newblock \emph{Navigating the social world: What infants, children, and other
  species can teach us}, pages 292--296.

\end{thebibliography}
\bibliographystyle{acl_natbib}

% \appendix

\end{document}